\documentclass[10pt,twocolumn,letterpaper]{article}

\usepackage{iccv}
\usepackage{times}
\usepackage{epsfig}
\usepackage{graphicx}
\usepackage{amsmath}
\usepackage{amssymb}

\usepackage{epsfig}
\usepackage{graphicx}
\usepackage{amsmath}
\usepackage{amssymb}
\usepackage{multirow}
\usepackage{enumerate}
\usepackage{bm}
\usepackage{wrapfig}
\usepackage{epstopdf}
\usepackage{wrapfig}

\usepackage[pagebackref=true,breaklinks=true,letterpaper=true,colorlinks,bookmarks=false]{hyperref}

\iccvfinalcopy 

\def\iccvPaperID{3404} 

\ificcvfinal\fi

\begin{document}

\title{GridPull: Towards Scalability in Learning Implicit Representations from 3D Point Clouds}

\author{Chao Chen$^1$ \quad
    Yu-Shen Liu$^1$
    \thanks{The corresponding author is Yu-Shen Liu. This work was supported by National Key R\&D Program of China (2022YFC3800600), the National Natural Science Foundation of China (62272263, 62072268), and in part by Tsinghua-Kuaishou Institute of Future Media Data.} \quad
    Zhizhong Han$^2$ \\
\fontsize{11pt}{\baselineskip}\selectfont{$^1$School of Software, Tsinghua University, Beijing, China} \\
\fontsize{11pt}{\baselineskip}\selectfont{$^2$Department of Computer Science, Wayne State University, Detroit, USA} \\
{\tt\small chenchao19@mails.tsinghua.edu.cn} \quad  {\tt\small liuyushen@tsinghua.edu.cn} \quad {\tt\small h312h@wayne.edu}
}

\maketitle
\ificcvfinal\thispagestyle{empty}\fi

\begin{abstract}
Learning implicit representations has been a widely used solution for surface reconstruction from 3D point clouds. The latest methods infer a distance or occupancy field by overfitting a neural network on a single point cloud. However, these methods suffer from a slow inference due to the slow convergence of neural networks and the extensive calculation of distances to surface points, which limits them to small scale points. To resolve the scalability issue in surface reconstruction, we propose GridPull to improve the efficiency of learning implicit representations from large scale point clouds. Our novelty lies in the fast inference of a discrete distance field defined on grids without using any neural components. To remedy the lack of continuousness brought by neural networks, we introduce a loss function to encourage continuous distances and consistent gradients in the field during pulling queries onto the surface in grids near to the surface. We use uniform grids for a fast grid search to localize sampled queries, and organize surface points in a tree structure to speed up the calculation of distances to the surface. We do not rely on learning priors or normal supervision during optimization, and achieve superiority over the latest methods in terms of complexity and accuracy. We evaluate our method on shape and scene benchmarks, and report numerical and visual comparisons with the latest methods to justify our effectiveness and superiority. The code is available at \url{https://github.com/chenchao15/GridPull}.
\end{abstract}

\section{Introduction}
It is vital to reconstruct surfaces from 3D point clouds for downstream applications. A widely used strategy is to learn implicit representations~\cite{mildenhall2020nerf,Oechsle2021ICCV,handrwr2020,zhizhongiccv2021matching,zhizhongiccv2021completing,takikawa2021nglod,DBLP:journals/corr/abs-2105-02788,rematasICML21,wen20223d} from 3D point clouds in a data-driven manner. With the learned implicit representations, we can reconstruct surfaces in meshes by running the marching cubes algorithm~\cite{Lorensen87marchingcubes}. By learning priors from large scale datasets during training, previous methods~\cite{Mi_2020_CVPR,Genova:2019:LST,jia2020learning,Liu2021MLS,tang2021sign,Peng2020ECCV,ErlerEtAl:Points2Surf:ECCV:2020} generalize the learned priors to either global implicit functions~\cite{Williams_2019_CVPR,Tretschk2020PatchNets,DBLP:conf/eccv/ChabraLISSLN20,jiang2020lig,Boulch_2022_CVPR,DBLP:conf/cvpr/MaLH22} or local ones for unseen point clouds. However, the generalization ability limits their performances on large structure or geometry variations of training samples.

More recent methods~\cite{DBLP:conf/icml/GroppYHAL20,chen2023unsupervised,Atzmon_2020_CVPR,zhao2020signagnostic,atzmon2020sald,yifan2020isopoints,DBLP:journals/corr/abs-2106-10811,Zhizhong2021icml,chibane2020neural,Peng2021SAP,jin2023multi} achieved better generalization by directly overfitting neural networks on single unseen point clouds. They rely on extensive calculations of distances between queries and surface points to probe the space, which supervises neural networks to converge to a distance or occupancy field~\cite{Zhizhong2021icml,DBLP:conf/icml/GroppYHAL20,Atzmon_2020_CVPR,zhao2020signagnostic,atzmon2020sald,chaompi2022,ma2023learning,Wang_2023_CVPR,DBLP:conf/cvpr/LiWLSH22}. However, this inference procedure is time consuming due to the extensive distance calculations and the slow convergence of neural networks. This demerit makes these methods hard to scale up to large scale point clouds for surface reconstruction.

To address the scalability challenge, we propose GridPull to speed up the learning of implicit function from large scale point clouds. GridPull does not require learned priors or point normal, and directly infers a distance field from a point cloud without using any neural components. We infer the distance field on grids near the surface, which reduces the number of grids we need to infer. Moreover, we organize surface points in a tree structure to speed up the nearest neighbor search for the calculation of distances to the surface. Specifically, we infer the discrete distance field by pulling queries onto the surface in grids of interests. Our loss function encourages continuous distances and consistent gradients in the field, which makes up the lack of continuousness brought by neural networks. We justify our effectiveness and highlight our superiority by numerical and visual comparisons with the latest methods on the widely used benchmarks. Our contributions are listed below.

\begin{enumerate}[i)]
\item We propose GridPull to reconstruct surfaces from large scale point clouds without using neural networks. GridPull speeds up the learning of implicit function, which addresses the scalability challenge in surface reconstruction.
\item We introduce a loss function to directly infer a discrete distance field defined on grids but achieving continuous distances and consistent gradients in the field.
\item Our method outperforms state-of-the-art methods in surface reconstruction in terms of speed and accuracy on the widely used benchmarks.
\end{enumerate}

\section{Related Work}
Neural implicit representations have made a huge progress in various tasks~\cite{mildenhall2020nerf,Oechsle2021ICCV,handrwr2020,zhizhongiccv2021matching,zhou2023levelset,zhizhongiccv2021completing,takikawa2021nglod,DBLP:journals/corr/abs-2105-02788,rematasICML21,ma2023towards,li2023neaf,jun2023shap,mueller2022instant,instructnerf2023,brooks2022instructpix2pix,li2023neuralangelo,sun2022improved}. We can use different supervision including 3D supervision~\cite{DBLP:journals/corr/abs-1901-06802,Park_2019_CVPR,aminie2022,MeschederNetworks,chen2018implicit_decoder,xiang2022snowflake,wen2022pmp}, multi-view~\cite{sitzmann2019srns,DIST2019SDFRcvpr,Jiang2019SDFDiffDRcvpr,prior2019SDFRcvpr,shichenNIPS,DBLP:journals/cgf/WuS20,Volumetric2019SDFRcvpr,lin2020sdfsrn,yariv2020multiview,yariv2021volume,geoneusfu,neuslingjie,Yu2022MonoSDF,yiqunhfSDF,Vicini2022sdf,wang2022neuris,Han2019ShapeCaptionerGC,goli2023nerf2nerf,hertz2022prompt,rosu2023permutosdf,DBLP:journals/corr/abs-2111-11215,jiangCQN2023,meng_2023_neat}, and point clouds~\cite{Williams_2019_CVPR,liu2020meshing,Mi_2020_CVPR,Genova:2019:LST} to learn neural implicit representations. In the following, we focus on reviewing works on surface reconstruction by learning implicit representations point clouds below.

\noindent\textbf{Learning with Priors. }With neural networks, one intuitive strategy is to use neural networks to learn priors from a training set and then generalize the learned priors to unseen samples. We can learn either global priors to map a shape level point cloud into a global implicit function~\cite{Mi_2020_CVPR,Genova:2019:LST,jia2020learning,Liu2021MLS,tang2021sign,Peng2020ECCV,ErlerEtAl:Points2Surf:ECCV:2020} or local priors for local implicit functions to represent parts or patches~\cite{Williams_2019_CVPR,Tretschk2020PatchNets,DBLP:conf/eccv/ChabraLISSLN20,jiang2020lig,Boulch_2022_CVPR,DBLP:conf/cvpr/MaLH22} which are further used to approximate a global implicit function.

These methods require large scale datasets to learn priors. However, the learned priors may not generalize well to unseen point clouds that have large geometric variations compared to training samples. Our method does not require priors, and falls into the following category.

\noindent\textbf{Learning with Overfitting. }To improve the generalization, we can learn implicit functions by overfitting neural networks on single point clouds. Methods using this strategy introduce novel constraints~\cite{DBLP:conf/icml/GroppYHAL20,Atzmon_2020_CVPR,zhao2020signagnostic,atzmon2020sald,yifan2020isopoints,DBLP:journals/corr/abs-2106-10811}, ways of leveraging gradients~\cite{Zhizhong2021icml,chibane2020neural}, differentiable poisson solver~\cite{Peng2021SAP} or specially designed priors~\cite{DBLP:conf/cvpr/MaLH22,DBLP:conf/cvpr/MaLZH22} to learn signed~\cite{Zhizhong2021icml,DBLP:conf/icml/GroppYHAL20,Atzmon_2020_CVPR,zhao2020signagnostic,atzmon2020sald,chaompi2022} or unsigned distance functions~\cite{chibane2020neural,zhou2022learning,Liu23NeUDF}.

These methods rely on neural networks to infer implicit functions without learning priors. However, the slow convergence of neural networks is a big limit for them to scale up to large scale point clouds. Our method addresses the scalability challenge by directly inferring a discrete distance field without using neural networks. Our novel loss leads to less complexity and higher accuracy than the overfitting based methods.

\noindent\textbf{Learning with Grids. }Learning implicit functions with grids has been used in prior-based methods~\cite{takikawa2021nglod,DBLP:journals/corr/abs-2105-02788,DBLP:conf/eccv/PengNMP020,tang2021octfield,Liu2021MLS,Jiang2019SDFDiffDRcvpr,Wang_dual2022}. Using neural networks, these methods learn features at vertices of grids and further map the interpolated features into a distance or occupancy field. Some methods~\cite{yu_and_fridovichkeil2021plenoxels,SunSC22,yu2021plenoctrees,DBLP:journals/corr/abs-2208-10925} inferred a discrete radiance field defined on grids to speed up the learning of a radiance field, while they cared more about the quality of synthesized views than the underlying geometry. Without using neural components, some methods~\cite{VisCovolume,Peng2021SAP} directly infer discrete implicit functions on grids. To make up the continuousness of neural networks, they employed different constraints, such as Possion equations or Viscosity priors, to pursue a continuous field from a single point cloud. However, these methods are slowly converged, which makes them hard to scale up to large scale point clouds. Although we also use grids to speed up the inference, we introduce a more efficient training strategy to decrease optimization space and more effective losses to increase inference efficiency.

\section{Method}
\begin{figure*}[tb]
  \centering
   \includegraphics[width=\linewidth]{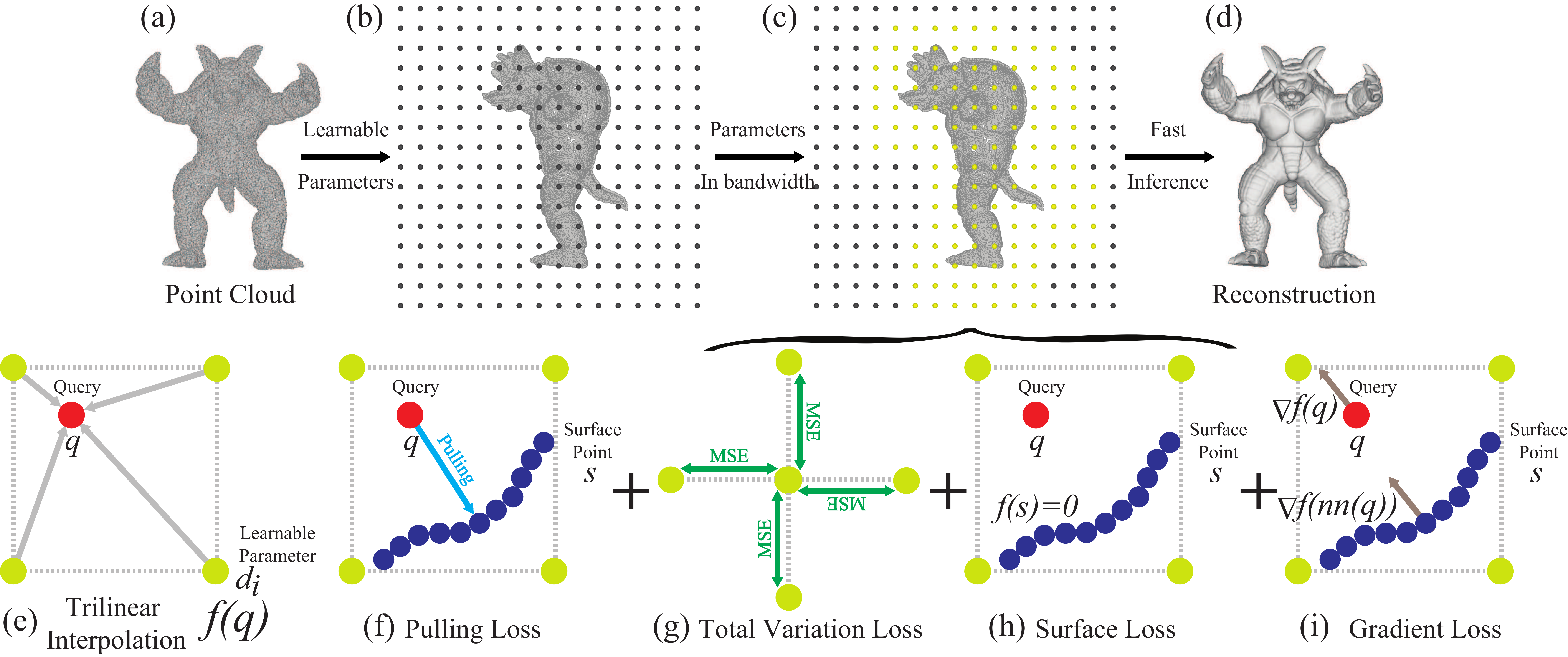}
  %
  %
  \vspace{-0.25in}
\caption{\label{fig:overview}We proposed GridPull for fast inference of SDF defined on discrete grids using effective losses.}
\vspace{-0.28in}
\end{figure*}

\noindent\textbf{Overview. }GridPull aims to achieve a fast inference of a distance field, such as a signed distance field, from a 3D point cloud $S=\{s_j|j\in[1,J]\}$ without using any neural components. As shown in Fig.~\ref{fig:overview} (b), we represent the distance field by a signed distance function (SDF) $f$ defined on a set of discrete grids $G$ with a resolution of $R^3$, where the vertices $V=\{v_i|i\in[1,I],I=(R+1)^3\}$ shared among these grids hold learnable signed distances $D=\{d_i\}$. For an arbitrary location $q$ in Fig.~\ref{fig:overview} (e), such as a randomly sampled query or a surface point, we use trilinear interpolation to approximate the signed distance $f(q)$ from the eight nearest grid vertices below,

\begin{equation}
\label{eq:sdf}
f(q)=triInter(q,D).
\end{equation}

Our goal is to conduct direct optimization in Fig.~\ref{fig:overview} (c) to infer the signed distances $D$ on grid vertices by minimizing our loss function $L$ on queries $q$ below,

\begin{equation}
\label{eq:objective}
\min_{D} \ L(f(q)).
\end{equation}

With the learned function $f$, we run the marching cubes algorithm~\cite{Lorensen87marchingcubes} to reconstruct a mesh as the surface as shown in Fig.~\ref{fig:overview} (d).

\noindent\textbf{Discrete Grids. }We split the space occupied by the point cloud $S$ into discrete grids $G$ with a resolution of $R$ in Fig.~\ref{fig:overview} (b). GridPull can work with high resolutions, such as $R=512$, because of our efficient and effective loss function and training strategy. Within each grid, we interpolate the signed distance $f(q)$ of query $q$ using the trilinear interpolation in Eq.~\ref{eq:sdf}. We organize surface points in a tree structure to speed up the search of the nearest surface point for queries $q$. Since we need to extensively conduct trilinear interpolation on signed distances for queries $q$, we use uniform grids for fast nearest grid search for queries. We did not use an Octree to organize grids, due to its inefficient and inaccurate grid search for queries.


\noindent\textbf{Geometric Initialization. }We initialize the signed distances $d_i$ on grid vertices which represent a sphere in the field. The initialization can speed up the convergence during optimization. Our preliminary results show that a small sphere with a radius of one or two grids works well. Since enlarging a small sphere to the target shape can calculate the continuous distance loss over smaller regions than shrinking a large sphere.

\noindent\textbf{Signed Distance Inference. }For a query $q$, we infer its signed distance $f(q)$ through pulling it onto its nearest surface point in the point cloud $S$. As shown in Fig.~\ref{fig:overview} (e), we first obtain $f(q)$ as the interpolation of the signed distances $d_i$ on the nearest eight vertices of $q$. We implement the trilinear interpolation by solving $\bm{a}=[a_0,..,a_7]$ in a linear regression system on the eight vertices $v_i$ below,

\vspace{-0.15in}
\begin{equation}
\label{eq:trilinear}
f(q)=a_0+a_1x+a_2y+a_3z+a_4xy+a_5xz+a_6yz+a_7xyz,
\end{equation}

\noindent where $[x,y,z]$ is the coordinate of $q$ and the eight constraints are $\{f(v_i)=d_i\}$.

With the predicted signed distance $f(q)$ and the gradient $\nabla f(q)=\partial f/\partial q$, we pull a query $q$ onto the surface and obtain a pulled query $p$ by $p=q-f(q)\times\nabla f(q)/||\nabla f(q)||_2$ in Fig.~\ref{fig:overview} (f).

Our goal is to minimize the distance between the pulled query $p$ and the nearest surface point $NN(q)\in S$ of query $q$ below,

\vspace{-0.15in}
\begin{equation}
\label{eq:infer}
\begin{aligned}
L_{Pull}=||p-NN(q)||_2.
\end{aligned}
\end{equation}


We adopt this pulling loss from~\cite{Zhizhong2021icml}. Instead, we do not rely on neural network in the pulling procedure. We use the trilinear interpolation on signed distances $d_i$ on grid vertices to obtain signed distances and gradients according to Eq.~\ref{eq:trilinear}.

We sample queries $q$ around surface points in each epoch. For each randomly sampled surface point, we sample queries using a gaussian distribution with the surface point as the center and a two-grid length variance. We build a KD-tree on surface points so that we can speed up the searching of the nearest neighbor from a large number of surface points for each query.

To speed up the optimization, we only calculate the pulling loss on grids near the surface. We regard the grids containing surface points as the base and find the union of their $M_1$ nearest neighboring grids. These grids form a $M_1$-bandwidth near the surface in Fig.~\ref{fig:overview} (c).

\begin{figure}[tb]
  \centering
   \includegraphics[width=\linewidth]{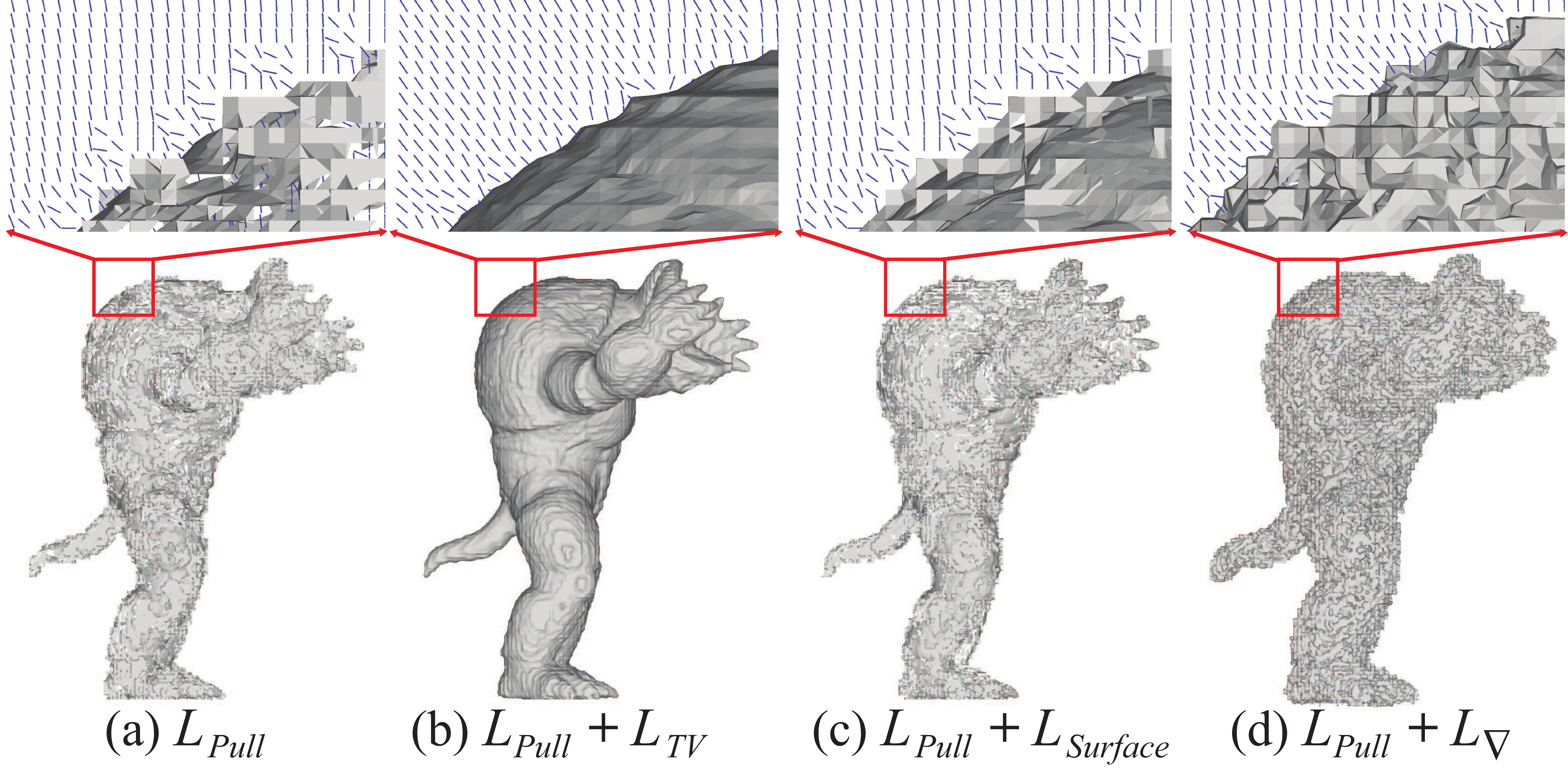}
  %
  %
  \vspace{-0.25in}
\caption{\label{fig:LossComp}The effect of losses on reconstruction.}
\vspace{-0.28in}
\end{figure}

\noindent\textbf{Continuous Distance Fields. }Although the pulling loss can infer correct signed distances at grid vertices, it fails to learn continuous signed distances across the field, especially near the surface. Fig.~\ref{fig:LossComp} (a) shows severe artifacts on the surface and gradients near the surface are messy and inconsistent, which are caused by the discontinuous signed distances across neighboring regions.

Neural network based methods~\cite{Zhizhong2021icml,DBLP:conf/cvpr/MaLZH22} do not have this issue, since the continuous character of neural networks makes the sudden change of signs on the same shape side hard. While our method uses signed distances $d_i$ on grid vertices to represent a discrete field, this makes that one $d_i$ can change separately without considering the change of neighboring $d_i$, which leads to a poor continuousness in the field.

To resolve this issue, we impose a total variation (TV) loss as a continuous constraint on our signed distances $d_i$ on grid vertices. Our key idea is to make the change of one $d_i$ affect the change of its neighboring $NN(d_i)$, which smoothes the local distance field and propagates local updates to the rest grids. Hence, as shown in a 2D case in Fig.~\ref{fig:overview} (g), we approximate one $d_i$ to each one of its six neighbors by minimizing the difference between each two below,

\vspace{-0.15in}
\begin{equation}
\label{eq:TV}
\begin{aligned}
L_{TV}=\sqrt{\sum\nolimits_{a=\{x,y,z\}}e_{a+1}(d_i)^2+e_{a-1}(d_i)^2},
\end{aligned}
\end{equation}

\noindent where $e_{a+1}(d_i)$ is the difference between $d_i$ and the signed distance along the $a$ axis, i.e., $x,y,z$ axis. Fig.~\ref{fig:LossComp} (b) shows that the continuous constraints can significantly improve the signed distance field in terms of continuousness and gradients.

One remaining issue is its computational complexity. Since we are targeting grids at a high resolution, such as $R=512$, it is very time consuming to impose this constraint on all $d_i$ across the field.

To resolve this issue, we focus on grids near the surface. Similar to imposing the pulling loss on grids in the $M_1$-bandwidth near the surface, we impose our continuous constraints on the grids in the $M_2$-bandwidth near the surface, which significantly reduces the number of learnable parameters to optimize.

\noindent\textbf{Signed Distance Supervision. }For signed distances, one supervision we can directly use is the input point cloud $S$. It is a discrete surface indicating the zero level-set of the signed distance function. Hence, as introduced in Fig.~\ref{fig:overview} (h), we can constrain the signed distances at the points $s$ on the surface below,

\vspace{-0.15in}
\begin{equation}
\label{eq:surface}
\begin{aligned}
L_{Surface}=||f(s)||_2.
\end{aligned}
\end{equation}

Although Fig.~\ref{fig:LossComp} (c) indicates that $L_{Surface}$ slightly improves the signed distance field and leads to a little bit more continuous surface, it is helpful to determine the zero level set with other losses. For noises on $S$, our continuous constraint can relieve the impact of noises on this supervision. We will show this in experiments.

\noindent\textbf{Consistent Gradients. }Fig.~\ref{fig:LossComp} (a) indicates that the pulling loss produces gradients that are messy and non-orthogonal to the surface. This leads to inaccurate signed distance inference during the pulling procedure.

We additionally introduce a constraint on gradients to improve the field near the surface. As shown in Fig.~\ref{fig:overview} (i), we aim to encourage the gradient at a query $q$ to point to the same direction of the gradient at its nearest surface point $NN(q)$. We use a cosine distance below to achieve more consistent gradients,
\vspace{-0.1in}
\begin{equation}
\label{eq:gradient}
\begin{aligned}
L_{\nabla}=1-cos(\nabla f(q),\nabla f(NN(q))).
\end{aligned}
\end{equation}

$L_{\nabla}$ constrains the gradients in the field and corrects inaccurate gradients. Fig.~\ref{fig:LossComp} (d) shows that more consistent gradients can improve the surface to be more continuous using small planes on the surface which however the surface not smooth at all.

\begin{figure*}[tb]
\vspace{-0.15in}
  \centering
   \includegraphics[width=\linewidth]{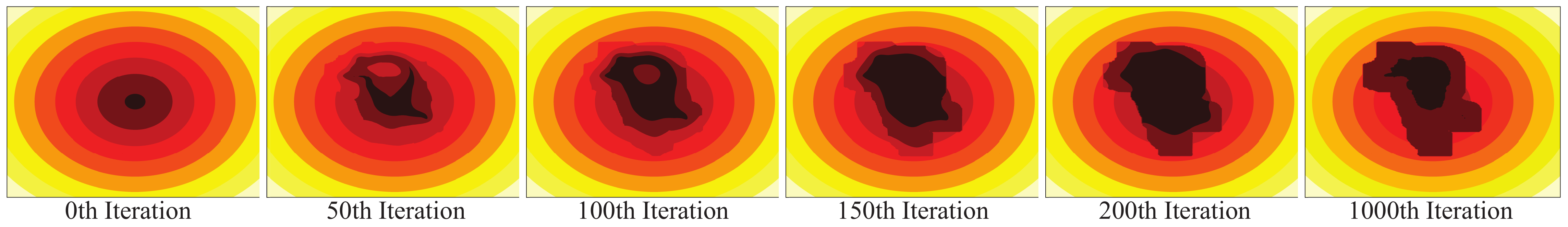}
  %
  %
  \vspace{-0.25in}
\caption{\label{fig:OptVis}Visualization of signed distances in optimization.}
\vspace{-0.25in}
\end{figure*}

\noindent\textbf{Loss Function. }We learn signed distances on grid vertices $D$ by minimizing the loss function below,

\vspace{-0.15in}
\begin{equation}
\label{eq:objective}
L=L_{Pull}+\alpha L_{TV}+\beta L_{Surface}+\gamma L_{\nabla}.
\end{equation}

\noindent where $\alpha$, $\beta$, and $\gamma$ are balance weights to make each term contribute equally. Fig.~\ref{fig:OptVis} shows the signed distance field on a slide of $D$ during optimization. Our optimization starts from an initial sphere and only optimizes variables in the bandwidth. We skip the grids that are not optimized after the initialization during the marching cubes for surface reconstruction. Please watch our video for more details.

\section{Experiments and Analysis}
We evaluate GridPull by numerical and visual comparisons with the latest methods on synthetic and real datasets in surface reconstruction.

\noindent\textbf{Datasets and Metrics. }We use benchmarks for shapes and scenes in evaluations. For shapes, we conduct evaluations on five datasets including a subset of ShapeNet~\cite{shapeneturl}, FAMOUS~\cite{ErlerEtAl:Points2Surf:ECCV:2020}, Thingi10k~\cite{DBLP:journals/corr/ZhouJ16}, Surface Reconstruction Benchmark (SRB)~\cite{Williams_2019_CVPR} and D-FAUST~\cite{dfaust:CVPR:2017}. For scenes, we also report comparisons on five datasets including 3DScene~\cite{DBLP:journals/tog/ZhouK13}, SceneNet~\cite{7780811Handa}, 3DFRONT~\cite{DBLP:journals/corr/abs-2011-09127}, Matterport~\cite{Matterport3D} and KITTI~\cite{Geiger2012CVPR}.

We use L2 Chamfer distance ($CD_{L2}$), $CD_{L1}$ and Hausdorff distance(HD) to measure the error between the reconstructed surface and the ground truth. Moreover, we use normal consistency (NC) and F-score to evaluate the accuracy of normal on the reconstructed surface. We also report our time and storage complexity to highlight our advantage towards scalability. We also use the intersection over union (IoU) to measure the reconstruction for fair comparison with the latest methods.

\noindent\textbf{Details. }We use $R=256$ to infer a distance field. We   calculate the pulling loss on grids in a bandwidth with $M_1=3$. We impose the continuous constraint on grids in a bandwidth with $M_2=14$ for shapes and $M_2=25$ for scenes, since scenes contain open surfaces with larger empty spaces. We run the marching cubes for surface reconstruction at a resolution of 256.

We use the Adam optimizer with an initial learning rate of 1.0. We decrease the learning rate with a decay rate of 0.3 every 400 iterations. We use $50,000$ queries in each iteration, and run 1600 iterations for each point cloud during optimization. We set the balance weights $\alpha=1$, $\beta=1$, and $\gamma=0.005$ for equal contribution from the three terms.

\begin{figure}[tb]
  \centering
   \includegraphics[width=\linewidth]{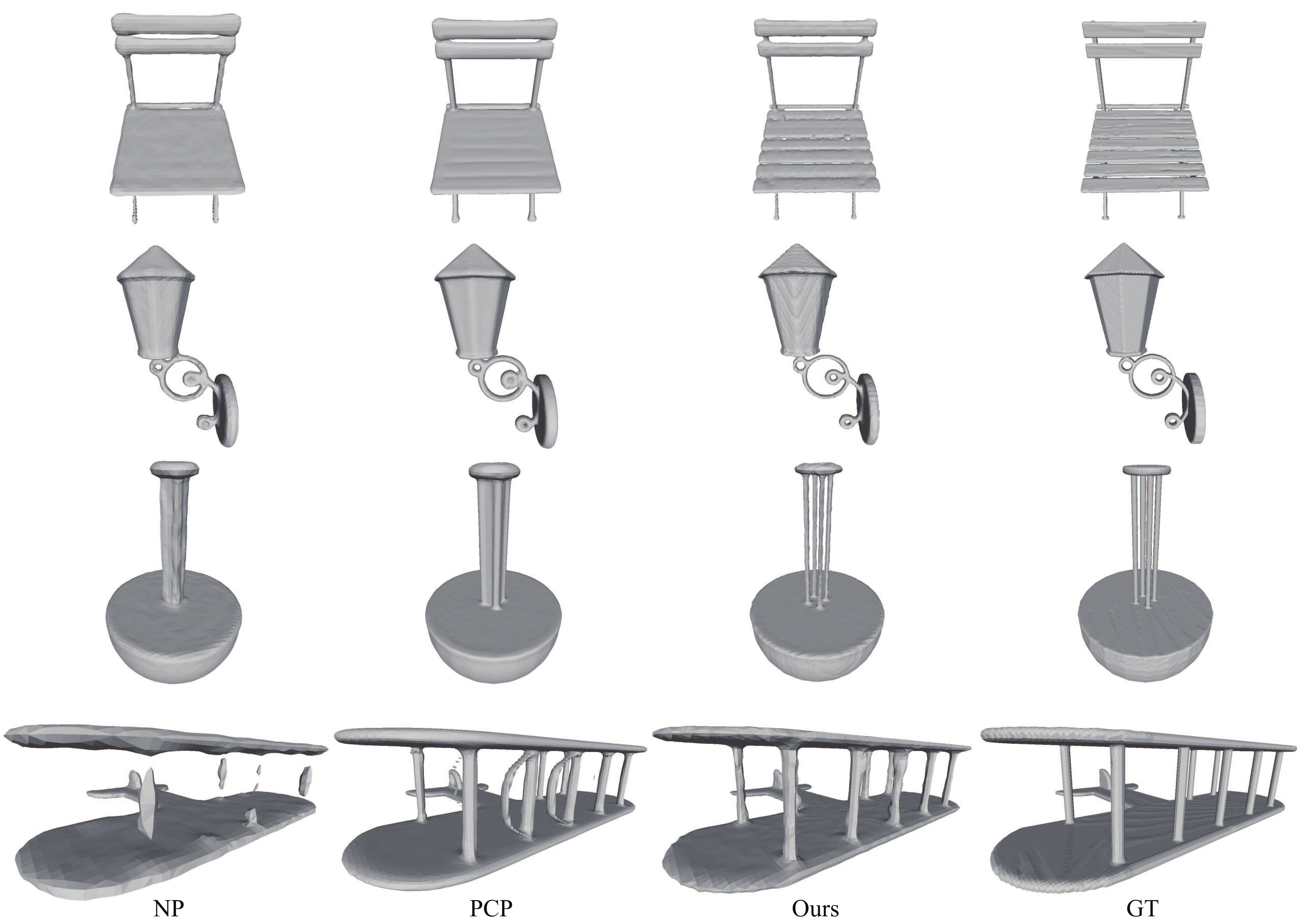}
  %
  %
  \vspace{-0.20in}
\caption{\label{fig:Shapetnet}Visual comparison under ShapetNet dataset.}
\vspace{-0.15in}
\end{figure}

\subsection{Surface Reconstruction for Shapes}

\noindent\textbf{ShapeNet. }Following predictive context prior (PCP)~\cite{DBLP:conf/cvpr/MaLZH22} and NeuralPull (NP)~\cite{Zhizhong2021icml}, we report our results on a subset in ShapeNet in terms of $CD_{L2}$ in Tab.~\ref{table:shapenetl2}, Normal Consistency (NC) in Tab.~\ref{table:shapenetnc}, and F-Score with thresholds of $0.002$ and $0.004$ in Tab.~\ref{table:fscore1} and Tab.~\ref{table:fscore2}, which show our superiority over PCP and NP in numerical comparisons over all classes, even though PCP employs a local shape prior learned from a large scale dataset. The visual comparison in Fig.~\ref{fig:Shapetnet} shows that we can reveal more accurate and detailed geometry.



\begin{table}[h]
\vspace{-0.1in}
\centering
\resizebox{0.7\linewidth}{!}{
    \begin{tabular}{c|c|c|c}
     \hline
     Class&NP~\cite{Zhizhong2021icml}&PCP~\cite{DBLP:conf/cvpr/MaLZH22}&Ours\\
     \hline
     Display&0.039&0.0087&\textbf{0.0082}\\
     Lamp&0.080&0.0380&\textbf{0.0347}\\
     Airplane&0.008&0.0065&\textbf{0.0007}\\
     Cabinet&0.026&0.0153&\textbf{0.0112}\\
     Vessel&0.022&0.0079&\textbf{0.0033}\\
     Table&0.060&0.0131&\textbf{0.0052}\\
     Chair&0.054&0.0110&\textbf{0.0043}\\
     Sofa&0.012&0.0086&\textbf{0.0015}\\
     \hline
     Mean&0.038&0.0136&\textbf{0.0086}\\
     \hline
   \end{tabular}}
    \vspace{0.05in}
   \caption{Reconstruction accuracy under ShapeNet in terms of $CD_{L2}\times100$.}
   \label{table:shapenetl2}
   \vspace{-0.1in}
\end{table}

\begin{table}[h]
\vspace{0.1in}
\centering
\resizebox{0.7\linewidth}{!}{
    \begin{tabular}{c|c|c|c}
     \hline
     Class&NP~\cite{Zhizhong2021icml}&PCP~\cite{DBLP:conf/cvpr/MaLZH22}&Ours\\
     \hline
     Display&0.964&0.9775&\textbf{0.9847}\\
     Lamp&0.930&0.9450&\textbf{0.9693}\\
     Airplane&0.947&0.9490&\textbf{0.9614}\\
     Cabinet&0.930&0.9600&\textbf{0.9689}\\
     Vessel&0.941&0.9546&\textbf{0.9667}\\
     Table&0.908&0.9595&\textbf{0.9755}\\
     Chair&0.937&0.9580&\textbf{0.9733}\\
     Sofa&0.951&0.9680&\textbf{0.9792}\\
     \hline
     Mean&0.939&0.9590&\textbf{0.9723}\\
     \hline
   \end{tabular}}
    \vspace{0.05in}
   \caption{Reconstruction accuracy under ShapeNet in terms of NC.}
   \label{table:shapenetnc}
   \vspace{-0.2in}
\end{table}

\begin{table}[h]
\vspace{0.1in}
\centering
\resizebox{0.7\linewidth}{!}{
    \begin{tabular}{c|c|c|c}
     \hline
     Class&NP~\cite{Zhizhong2021icml}&PCP~\cite{DBLP:conf/cvpr/MaLZH22}&Ours\\
     \hline
     Display&0.989&0.9939&\textbf{0.9963}\\
     Lamp&0.891&0.9382&\textbf{0.9455}\\
     Airplane&0.996&0.9942&\textbf{0.9976}\\
     Cabinet&0.980&0.9888&\textbf{0.9901}\\
     Vessel&0.985&0.9935&\textbf{0.9956}\\
     Table&0.922&0.9969&\textbf{0.9977}\\
     Chair&0.954&0.9970&\textbf{0.9979}\\
     Sofa&0.968&0.9943&\textbf{0.9974}\\
     \hline
     Mean&0.961&0.9871&\textbf{0.9896}\\
     \hline
   \end{tabular}}
    \vspace{0.05in}
   \caption{Reconstruction accuracy under ShapeNet in terms of F-Score with a threshold of 0.002.}
   \label{table:fscore1}
   \vspace{-0.0in}
\end{table}

\begin{table}[h]
\vspace{-0.1in}
\centering
\resizebox{0.7\linewidth}{!}{
    \begin{tabular}{c|c|c|c}
     \hline
     Class&NP~\cite{Zhizhong2021icml}&PCP~\cite{DBLP:conf/cvpr/MaLZH22}&Ours\\
     \hline
     Display&0.991&0.9958&\textbf{0.9963}\\
     Lamp&0.924&0.9402&\textbf{0.9538}\\
     Airplane&0.997&0.9972&\textbf{0.9989}\\
     Cabinet&0.989&0.9939&\textbf{0.9946}\\
     Vessel&0.990&0.9958&\textbf{0.9972}\\
     Table&0.973&0.9985&\textbf{0.9990}\\
     Chair&0.969&\textbf{0.9991}&0.9990\\
     Sofa&0.974&0.9987&\textbf{0.9992}\\
     Mean&0.976&0.9899&\textbf{0.9923}\\
     \hline
   \end{tabular}}
    \vspace{0.05in}
   \caption{Reconstruction accuracy under ShapeNet in terms of F-Score with a threshold of 0.004.}
   \label{table:fscore2}
   \vspace{-0.2in}
\end{table}

\begin{figure}[tb]
  \centering
   \includegraphics[width=\linewidth]{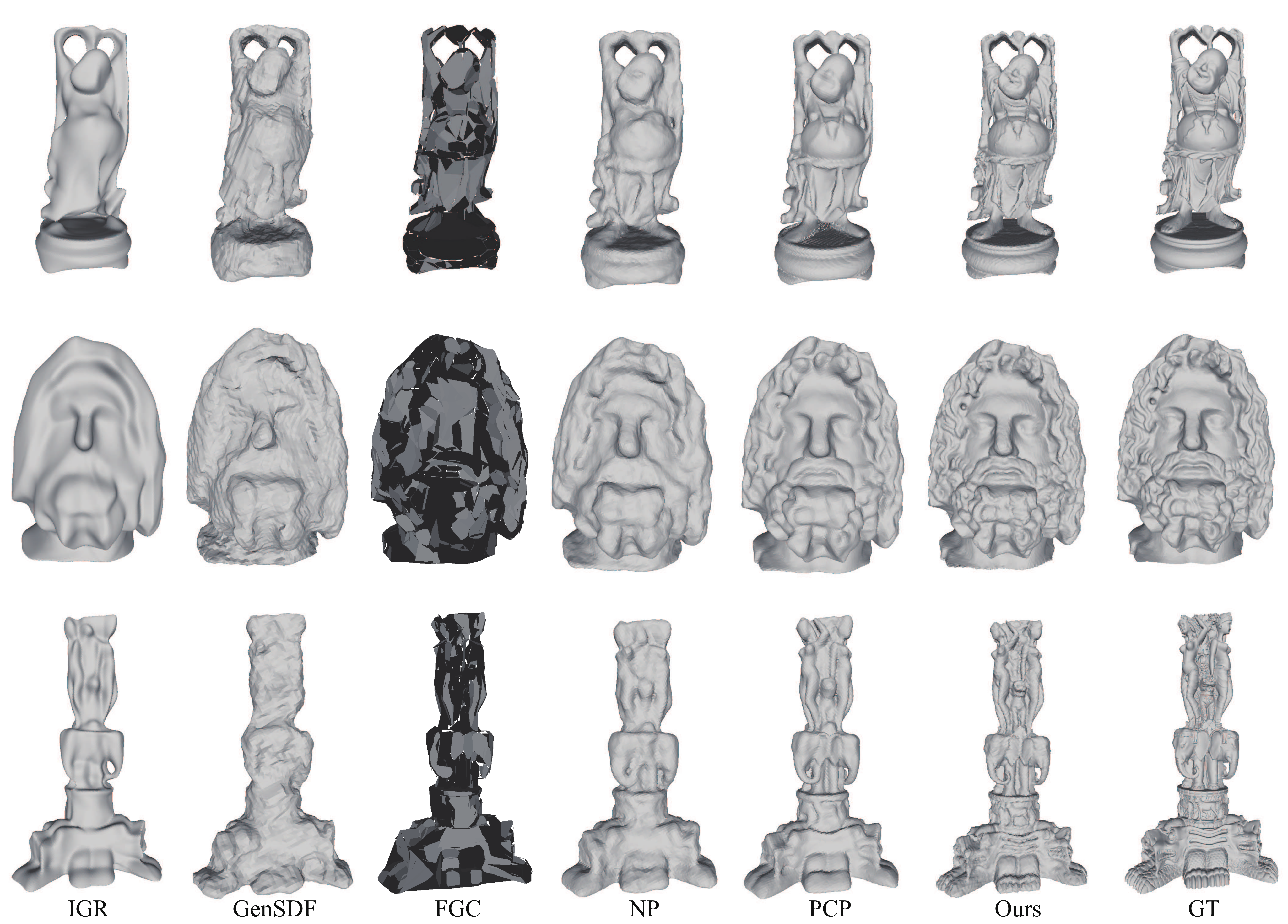}
  %
  %
  \vspace{-0.2in}
\caption{\label{fig:Famous}Visual comparison under FAMOUS dataset.}
\vspace{-0.1in}
\end{figure}

\noindent\textbf{FAMOUS. }We follow the experimental setting in PCP~\cite{DBLP:conf/cvpr/MaLZH22} and NP~\cite{Zhizhong2021icml} to evaluate GridPull on FAMOUS. We compare GridPull with the latest methods with priors including PCP, GenSDF, FGC and no priors including NP and IGR. GridPull outperforms these methods in terms of $CD_{L2}$ and running time. Although FGC is fast, it does not reconstruct a watertight mesh but just a polygon soup with no normal, which contains almost no geometry details. Worse than that, it fails to reconstruct some point clouds in more than eight hours, hence we remove these shapes from its results. We highlight our superiority in visual comparison in Fig.~\ref{fig:Famous}. We can reveal details like hair and expressions more clearly.

\begin{table}[h]
\vspace{-0.0in}
\centering
\resizebox{0.8\linewidth}{!}{
    \begin{tabular}{c|c|c|c}
     \hline
     Method&$CD_{L2}\times100$&Min time&Mean time\\
     \hline
     IGR~\cite{gropp2020implicit}&1.650&402s&407s\\
     GenSDF~\cite{chou2022gensdf}&0.668&230s&232s\\
     NP~\cite{Zhizhong2021icml}&0.220&593s&605s\\
     FGC~\cite{Yu_2022_CVPR}&0.055&10s&Timeout\\
     PCP~\cite{DBLP:conf/cvpr/MaLZH22}&0.044&3416s&3534s\\
     \hline
    Ours&\textbf{0.040}&198s&201s\\
     \hline
   \end{tabular}}
    \vspace{0.05in}
   \caption{Reconstruction accuracy under FAMOUS.}
   \label{table:famous}
   \vspace{-0.1in}
\end{table}

\begin{figure}[tb]
  \centering
   \includegraphics[width=\linewidth]{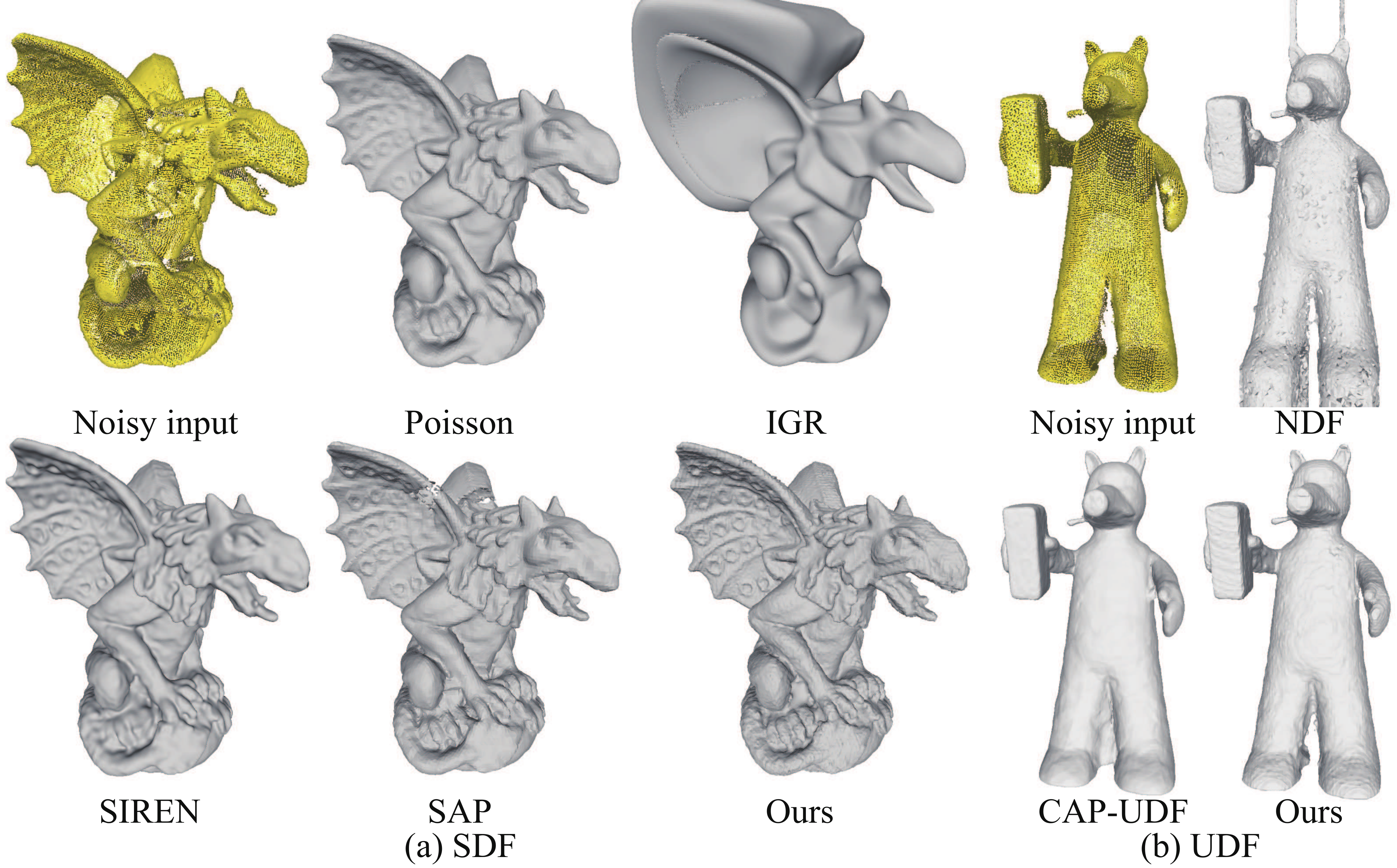}
  %
  %
  \vspace{-0.2in}
\caption{\label{fig:Srb}Visual comparison under SRB dataset.}
\vspace{-0.28in}
\end{figure}

\noindent\textbf{SRB. }We follow the experimental setting in VisCo~\cite{VisCovolume} to evaluate GridPull on real scans in SRB. We report the comparison with the latest overfitting based methods in terms of $CD_{L1}$ in Tab.~\ref{table:srbfull}. Meanwhile, we also report their one-sided distances between the reconstructed mesh and the input noisy point cloud. The numerical comparison shows that our method significantly outperforms the latest. Moreover, we also report our results of learning unsigned distances in Tab.~\ref{table:srbudf}. We replace our pulling loss with the pulling loss introduced for unsigned distances in CAP-UDF~\cite{zhou2022learning}, and remove the gradient loss due to the nondifferentiable character of UDF on surface. Visual comparison in Fig.~\ref{fig:Srb} shows that our method can reconstruct surfaces that are more compact than the methods learning SDF like SIREN or more smooth than the methods learning UDF like CAP-UDF.


\begin{table}[h]
\vspace{0.20in}
\centering
\resizebox{\linewidth}{!}{
    \begin{tabular}{c|c|c|c|c|c|c|c}
     \hline    &&Poisson~\cite{journals/tog/KazhdanH13}&IGR~\cite{gropp2020implicit}&SIREN~\cite{sitzmann2019siren}&VisCo~\cite{VisCovolume}&SAP~\cite{Peng2021SAP}&Ours\\
     \hline
     \multirow{4}{*}{Anchor}&$CD_{L1}$&0.60&0.22&0.32&0.21&0.12&\textbf{0.093}\\
     &$HD$&14.89&4.71&8.19&3.00&2.38&\textbf{1.804}\\
     &$d_{\overrightarrow C}$&0.60&0.12&0.10&0.15&0.08&\textbf{0.066}\\
     &$d_{\overrightarrow H}$&14.89&1.32&2.43&1.07&0.83&\textbf{0.460}\\
     \hline
     \multirow{4}{*}{Daratech}&$CD_{L1}$&0.44&0.25&0.21&0.26&0.07&\textbf{0.062}\\
     &$HD$&7.24&4.01&4.30&4.06&0.87&\textbf{0.648}\\
     &$d_{\overrightarrow C}$&0.44&0.08&0.09&0.14&0.04&\textbf{0.039}\\
     &$d_{\overrightarrow H}$&7.24&1.59&1.77&1.76&0.41&\textbf{0.293}\\
     \hline
     \multirow{4}{*}{DC}&$CD_{L1}$&0.27&0.17&0.15&0.15&0.07&\textbf{0.066}\\
     &$HD$&3.10&2.22&2.18&2.22&1.17&\textbf{1.103}\\
     &$d_{\overrightarrow C}$&0.27&0.09&0.06&0.09&0.04&\textbf{0.036}\\
     &$d_{\overrightarrow H}$&3.10&2.61&2.76&2.76&\textbf{0.53}&0.539\\
     \hline
     \multirow{4}{*}{Gargoyle}&$CD_{L1}$&0.26&0.16&0.17&0.17&0.07&\textbf{0.063}\\
     &$HD$&6.80&3.52&4.64&4.40&1.49&\textbf{1.129}\\
     &$d_{\overrightarrow C}$&0.26&0.06&0.08&0.11&0.05&\textbf{0.045}\\
     &$d_{\overrightarrow H}$&6.80&0.81&0.91&0.96&0.78&\textbf{0.700}\\
     \hline
     \multirow{4}{*}{Lord Quas}&$CD_{L1}$&0.20&0.12&0.17&0.12&0.05&\textbf{0.047}\\
     &$HD$&4.61&1.17&0.82&1.06&0.98&\textbf{0.569}\\
     &$d_{\overrightarrow C}$&0.20&0.07&0.12&0.07&0.04&\textbf{0.031}\\
     &$d_{\overrightarrow H}$&4.61&0.98&0.76&0.64&0.51&\textbf{0.370}\\
     \hline
   \end{tabular}}
   \vspace{-0.05in}
   \caption{Reconstruction accuracy under SRB.}  
   \label{table:srbfull}
\end{table}


\begin{table}[h]
\vspace{-0.1in}
\centering
\resizebox{0.7\linewidth}{!}{
    \begin{tabular}{c|c|c|c}
     \hline
     &NDF~\cite{chibane2020neural}&CAP-UDF~\cite{zhou2022learning}&Ours(UDF)\\
     \hline
     $CD_{L1}$&0.238&0.073&\textbf{0.070}\\
     F-Score&68.6&84.5&\textbf{85.1}\\
     \hline
   \end{tabular}}
    \vspace{0.05in}
   \caption{Accuracy of reconstruction with UDF under SRB.}
   \label{table:srbudf}
   \vspace{-0.2in}
\end{table}

\begin{figure}[tb]
  \centering
   \includegraphics[width=\linewidth]{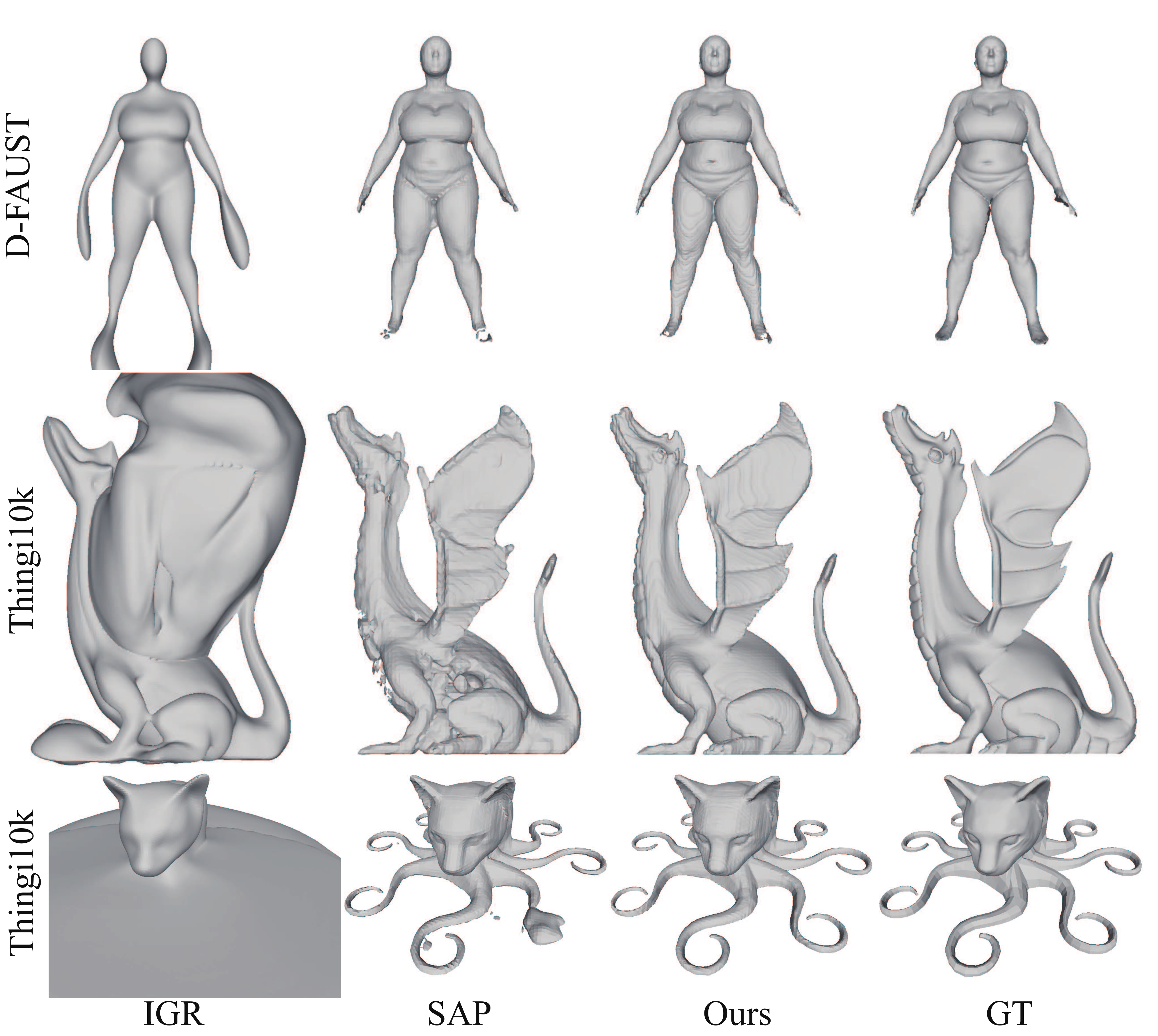}
  %
  %
  \vspace{-0.20in}
\caption{\label{fig:Dfaust_thingi10k}Visual comparison under Thingi10k and D-FAUST.}
\vspace{-0.05in}
\end{figure}

\noindent\textbf{Thingi10K. }We follow the experimental setting in SAP~\cite{Peng2021SAP} to evaluate GridPull in Thingi10K. Tab.~\ref{table:Thing10} indicates our superiority over the overfitting based methods in the numerical comparison. Visual comparisons in Fig.~\ref{fig:Dfaust_thingi10k} show our more accurate surfaces with complex details.

\begin{table}[h]
\vspace{-0.0in}
\centering
\resizebox{0.7\linewidth}{!}{
    \begin{tabular}{c|c|c|c|c}  
     \hline
     &&IGR~\cite{gropp2020implicit}&SAP~\cite{Peng2021SAP}&Ours\\
     \hline
     \multirow{3}{*}{Thing10k}&$CD_{L1}$&0.440&0.054&\textbf{0.051}\\
     &F-Score&0.505&0.940&\textbf{0.948}\\
     &NC&0.692&0.947&\textbf{0.965}\\
     \hline
     \multirow{3}{*}{DFAUST}&$CD_{L1}$&0.235&0.043&\textbf{0.015}\\
     &F-Score&0.805&0.966&\textbf{0.975}\\
     &NC&0.911&0.959&\textbf{0.978}\\
     \hline
   \end{tabular}}
   \vspace{0.05in}
   \caption{Reconstruction accuracy under Thingi10k and D-FAUST in terms of $CD_{L1}$ and F-Score with a threshold of 0.01.}
   \label{table:Thing10}
   \vspace{-0.1in}
\end{table}

\noindent\textbf{D-FAUST. }We also follow the experimental setting in SAP~\cite{Peng2021SAP} to evaluate GridPull in D-FAUST. Tab.~\ref{table:Thing10} shows that we outperform the overfitting based methods with more surface details. As shown in Fig.~\ref{fig:Dfaust_thingi10k}, we recover more accurate human bodies and more detailed expressions on faces.

\begin{figure*}[tb]
\vspace{-0.1in}
  \centering
   \includegraphics[width=\linewidth]{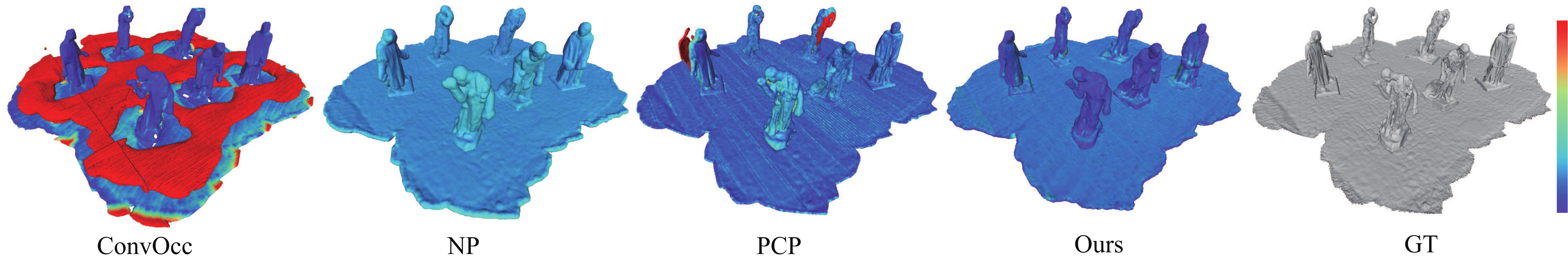}
  %
  %
  \vspace{-0.25in}
\caption{\label{fig:3dscene}Visual comparison under 3DScene dataset.}
\vspace{-0.15in}
\end{figure*}

\subsection{Surface Reconstruction for Scenes}
\noindent\textbf{3DScene. }We follow PCP~\cite{DBLP:conf/cvpr/MaLZH22} to report $CD_{L1}$, $CD_{L2}$ and Normal Consistency(NC) for evaluation. We report the comparisons with the latest methods in Tab.~\ref{table:3DScene}. We outperform both kinds of methods with learned priors such as ConvOcc~\cite{Peng2020ECCV} and PCP~\cite{DBLP:conf/cvpr/MaLZH22} and overfitting based NP~\cite{Zhizhong2021icml} in all scenes. The visual comparisons~\ref{fig:3dscene} shows that our method can work well with real scans on scenes and reveal more geometry details on surfaces.


 \begin{table}[h]
 \vspace{-0.1in}
 \centering
 \resizebox{\linewidth}{!}{
     \begin{tabular}{c|c|c|c|c|c}  
      \hline
      &&ConvOcc~\cite{Peng2020ECCV}&NP~\cite{Zhizhong2021icml}&PCP\cite{DBLP:conf/cvpr/MaLZH22}&Ours\\
      \hline
      \multirow{3}{*}{Burghers}&$CD_{L2}\times 100$&26.69&1.76&0.267&\textbf{0.246}\\
      &$CD_{L1}$&0.077&0.010&\textbf{0.008}&\textbf{0.008}\\
      &NC&0.865&0.883&0.914&\textbf{0.926}\\
      \hline
      \multirow{3}{*}{Lounge}&$CD_{L2}\times 100$&8.68&39.71&0.061&\textbf{0.055}\\      &$CD_{L1}$&0.042&0.059&0.006&\textbf{0.005}\\
      &NC&0.857&0.857&\textbf{0.928}&0.922\\
      \hline
      \multirow{3}{*}{Copyroom}&$CD_{L2}\times 100$&10.99&0.51&0.076&\textbf{0.069}\\
      &$CD_{L1}$&0.045&0.011&0.007&\textbf{0.006}\\
      &NC&0.848&0.884&0.918&\textbf{0.929}\\
      \hline
      \multirow{3}{*}{Stonewall}&$CD_{L2}\times 100$&19.12&0.063&0.061&\textbf{0.058}\\
      &$CD_{L1}$&0.066&0.007&0.0065&\textbf{0.006}\\
      &NC&0.866&0.868&0.888&\textbf{0.893}\\
      \hline
      \multirow{3}{*}{Totepole}&$CD_{L2}\times 100$&1.16&0.19&0.10&\textbf{0.093}\\
      &$CD_{L1}$&0.016&0.010&0.008&\textbf{0.007}\\
      &NC&0.925&0.765&0.784&\textbf{0.847}\\
      \hline
    \end{tabular}}
    \vspace{-0.0in}
    \caption{Reconstruction accuracy under 3DScene.}
    \label{table:3DScene}
    \vspace{-0.1in}
 \end{table}

\begin{figure}[tb]
  \centering
   \includegraphics[width=\linewidth]{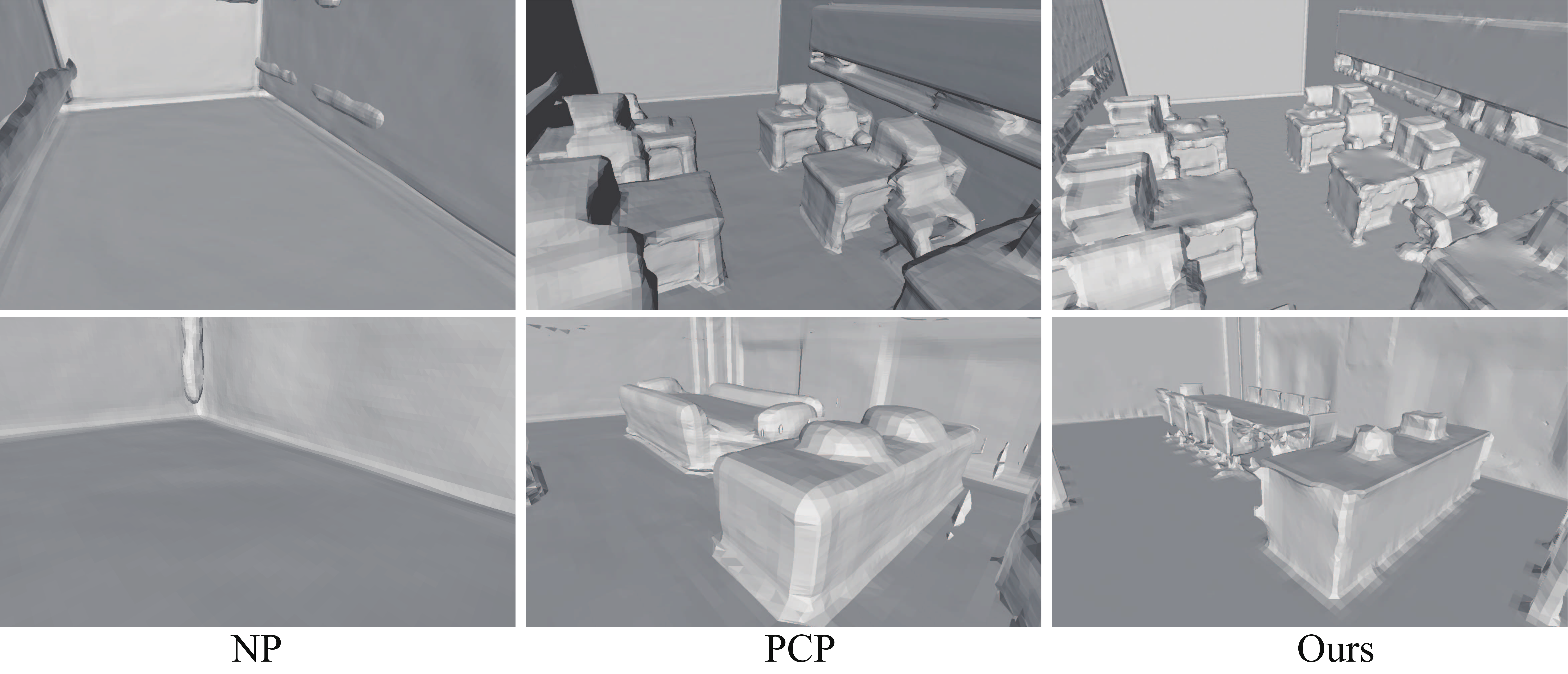}
  %
  %
  \vspace{-0.25in}
\caption{\label{fig:Scenenet}Visual comparison under SceneNet dataset.}
\vspace{-0.10in}
\end{figure}

\noindent\textbf{SceneNet. }Following the experimental setting in PCP~\cite{DBLP:conf/cvpr/MaLZH22}, we report our $CD_{L1}$, NC, and F-score in Tab.~\ref{table:SceneNet}. The comparison indicates that we outperform these methods by producing more accurate and smooth surfaces. This is also verified by our visual comparison in Fig.~\ref{fig:Scenenet}, where the latest methods reveal either no or inaccurate indoor geometry.


 \begin{table}[h]
 \vspace{0.0in}
 \centering
 \resizebox{0.8\linewidth}{!}{
     \begin{tabular}{c|c|c|c|c}  
      \hline
      &&NP~\cite{Zhizhong2021icml}&PCP~\cite{DBLP:conf/cvpr/MaLZH22}&Ours\\
      \hline
      \multirow{3}{*}{Livingroom}&$CD_{L1}$&0.088&0.017&\textbf{0.015}\\
      &$NC$&0.881&\textbf{0.933}&0.922\\
      &FScore&0.801&\textbf{0.966}&0.953\\
      \hline
      \multirow{3}{*}{Bathroom}&$CD_{L1}$&0.036&0.016&\textbf{0.013}\\
      &$NC$&0.912&0.945&\textbf{0.952}\\
      &FScore&0.860&\textbf{0.977}&0.974\\
      \hline
      \multirow{3}{*}{Bedroom}&$CD_{L1}$&0.034&\textbf{0.014}&\textbf{0.014}\\
      &$NC$&0.905&0.948&\textbf{0.951}\\
      &FScore&0.876&0.980&\textbf{0.986}\\
      \hline
      \multirow{3}{*}{Kitchen}&$CD_{L1}$&0.049&0.015&\textbf{0.014}\\
      &$NC$&0.900&0.945&\textbf{0.952}\\
      &FScore&0.825&0.976&\textbf{0.984}\\
      \hline
      \multirow{3}{*}{Office}&$CD_{L1}$&0.062&0.024&\textbf{0.022}\\
      &$NC$&0.879&0.919&\textbf{0.931}\\
      &FScore&0.729&925&\textbf{0.940}\\
      \hline
      \multirow{3}{*}{Mean}&$CD_{L1}$&0.054&0.017&\textbf{0.0156}\\
      &$NC$&0.895&0.938&\textbf{0.9416}\\
      &FScore&0.818&965&\textbf{0.9674}\\
      \hline
    \end{tabular}}
    \vspace{0.05in}
    \caption{Reconstruction accuracy under SceneNet. FScore is with a threshold of 0.025.}
    \label{table:SceneNet}
    \vspace{-0.0in}
 \end{table}

\begin{figure}[tb]
\vspace{-0.1in}
  \centering
   \includegraphics[width=\linewidth]{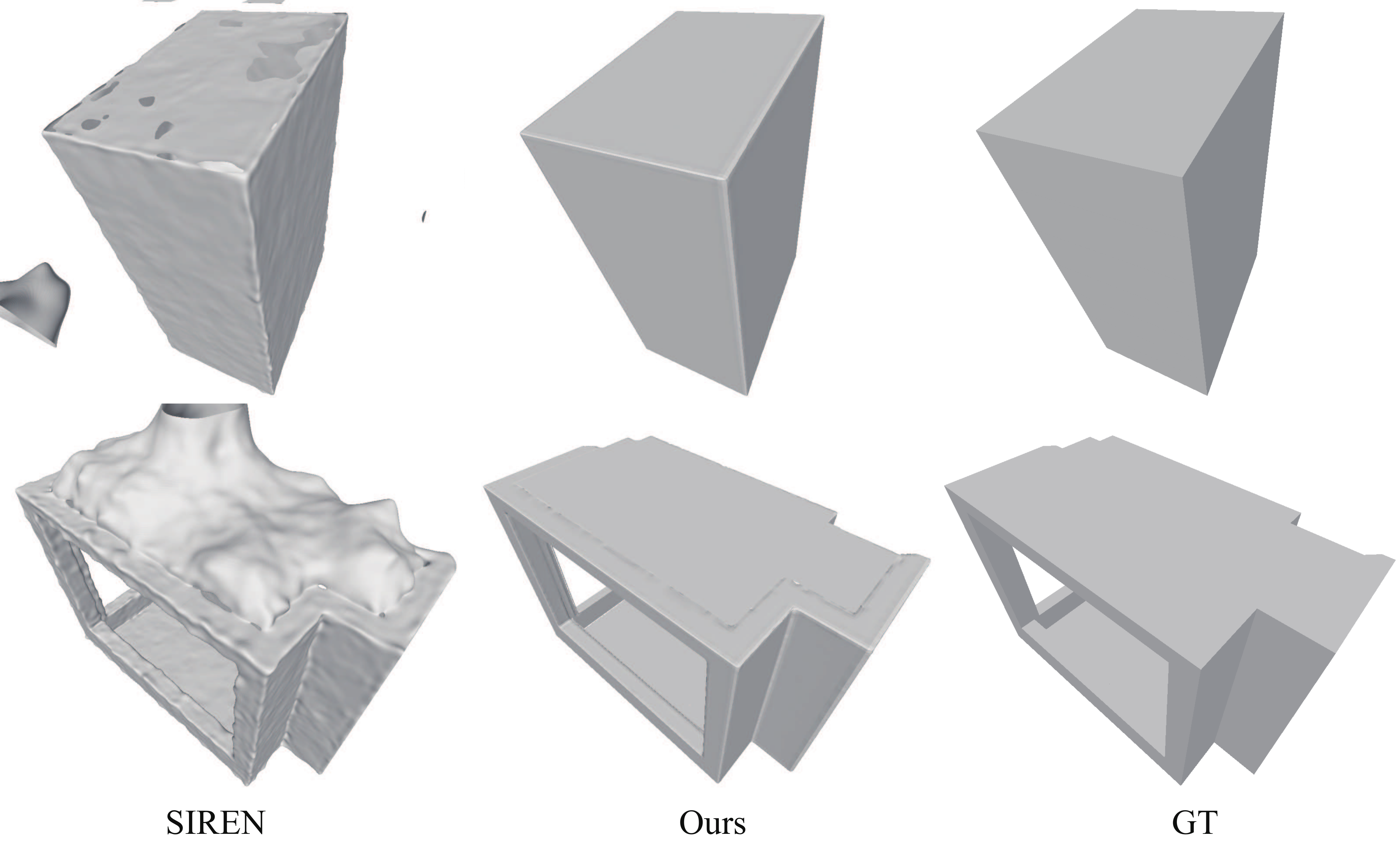}
  %
  %
  \vspace{-0.2in}
\caption{\label{fig:3dfront}Visual comparison under 3Dfront dataset.}
\vspace{-0.0in}
\end{figure}

\noindent\textbf{3D-FRONT. } We follow the experimental setting in NeuralPoisson~\cite{NeuralPoisson} to report $CD_{L1}$, IoU and L2 distance over the voxelization of meshes. IoU and L2 distances are calculated as the error between the reconstructed surfaces and the ground truth over the voxelization at a resolution of 128. The comparison in Tab.~\ref{table:3DFRONT} shows that our results are more accurate than the latest method. Our visual comparison with SIREN in Fig.~\ref{fig:3dfront} shows that GridPull can infer an accurate zero level set which leads to very sharp corners.

\begin{table}[h]
\vspace{-0.in}
\centering
\resizebox{0.6\linewidth}{!}{
    \begin{tabular}{c|c|c|c}
     \hline
     Method&CD&IOU&L2\\
     \hline
     SIREN~\cite{sitzmann2019siren}&0.0183&0.524&2.373\\
     NeuralPoisson~\cite{NeuralPoisson}&0.0161&0.545&2.073\\
     \hline
     Ours&\textbf{0.0141}&\textbf{0.562}&\textbf{1.984}\\
     \hline
   \end{tabular}}
    \vspace{0.05in}
   \caption{Reconstruction accuracy under 3DFRONT.}
   \label{table:3DFRONT}
   \vspace{-0.1in}
\end{table}

\begin{figure*}[tb]
\vspace{-0.0in}
  \centering
   \includegraphics[width=\linewidth]{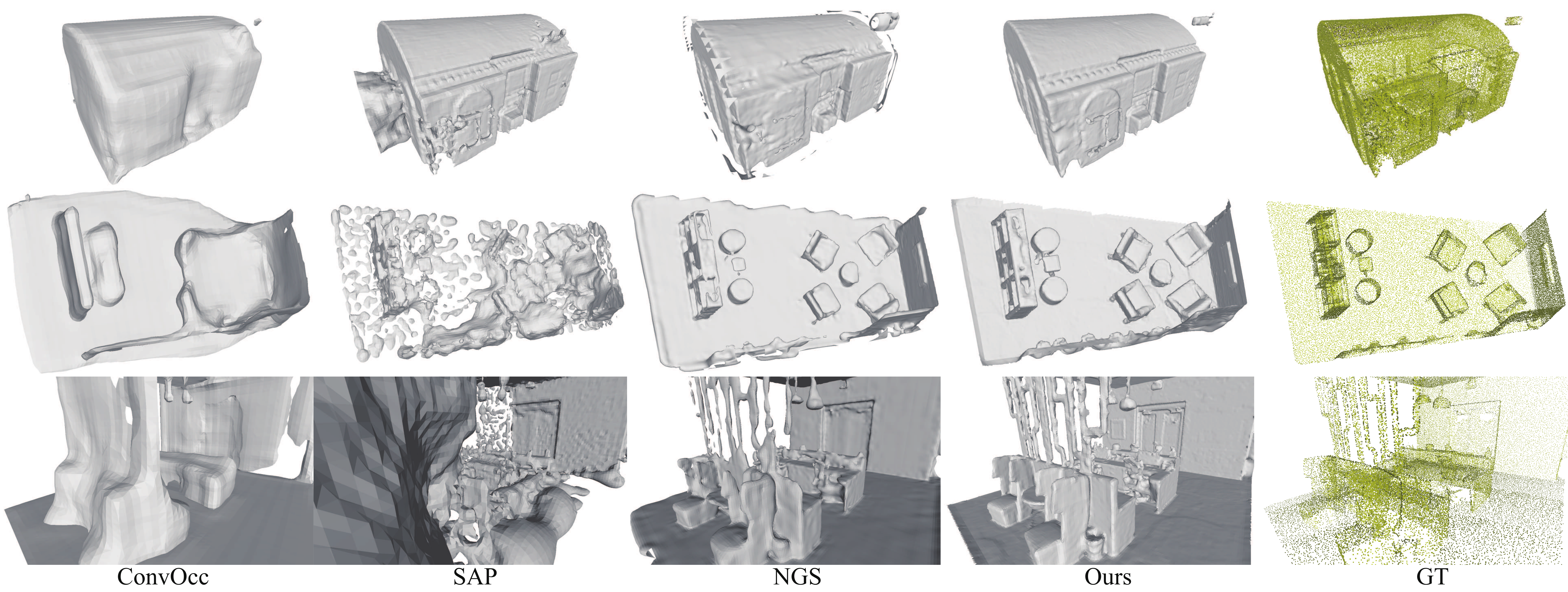}
  %
  %
  \vspace{-0.2in}
\caption{\label{fig:Matterport}Visual comparison under Matterport3D dataset.}
\vspace{-0.1in}
\end{figure*}

\noindent\textbf{Matterport. }We follow the experimental setting in NGS~\cite{huang2022neuralgalerkin}, and report the comparison in terms of $CD_{L1}$, NC, and F-score in Tab.~\ref{table:MATTER}. We achieve the best performance in terms of accuracy. NGS~\cite{huang2022neuralgalerkin} has a pretty fast inference but it costs about four to five days to learn priors from the training set in Matterport. Visual comparison in Fig.~\ref{fig:Matterport} shows that we can reconstruct more accurate surfaces while NGS is poorly generalized to unseen scenes.

\begin{figure}[tb]
\vspace{-0.0in}
  \centering
   \includegraphics[width=\linewidth]{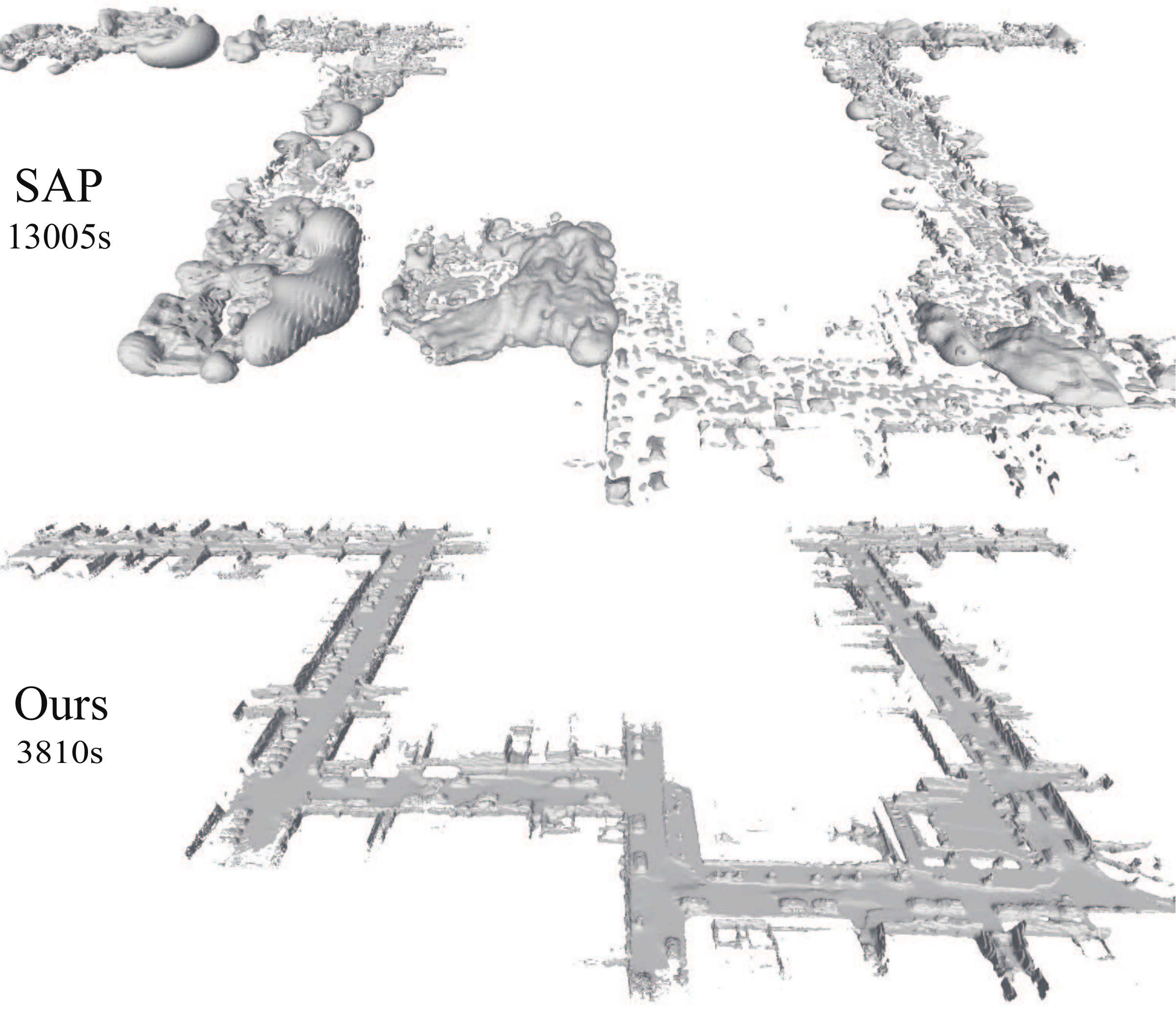}
  %
  %
  \vspace{-0.15in}
\caption{\label{fig:Kitti}Visual comparison under Kitti dataset.}
\vspace{-0.2in}
\end{figure}

\noindent\textbf{KITTI. }We evaluate GridPull on a large scale real scan from KITTI odometry (Sequence00, frame 3000 to 4000) which contains about $13.8$ million points. The comparison in Fig.~\ref{fig:Kitti} shows that we can reconstruct a more accurate surface in much less time ($3810$s) than the overfitting based method SAP ($13005$s)~\cite{Peng2021SAP}.


\begin{table}[h]
\vspace{0.0in}
\centering
\resizebox{0.9\linewidth}{!}{
    \begin{tabular}{c|c|c|c|c}
     \hline
     Method&$CD_{L1}\times1000$&F-Score(0.002)&NC&Time\\
     \hline
     ConvOcc~\cite{Peng2020ECCV}&6.21&88.9&92.0&34s\\
     SAP~\cite{Peng2021SAP}&4.17&91.2&91.4&643s\\
     NGS~\cite{huang2022neuralgalerkin}&3.04&97.3&95.7&2.1s\\
     \hline
     Ours&\textbf{2.93}&\textbf{97.7}&\textbf{96.3}&223s\\
     \hline
   \end{tabular}}
    \vspace{0.05in}
   \caption{Reconstruction accuracy under Matterport3D.}
   \label{table:MATTER}
   \vspace{-0.2in}
\end{table}

\subsection{Ablation Studies}
We conduct ablation studies on FAMOUS~\cite{ErlerEtAl:Points2Surf:ECCV:2020} to justify the effectiveness of modules in GridPull.

\noindent\textbf{Loss. }We justify the effectiveness of our loss in terms of $CD_{L2}$ in Tab.~\ref{table:ASLOSS}. For the pulling loss, it is able to estimate a coarse shape but with discontinuous meshes and artifacts. The gradient loss can improve the continuousness on the surface. Using the TV loss, we obtained a more continuous field, which removes most of artifacts and significantly improve the accuracy. The surface loss encourages a more accurate zero level-set, which also improves the accuracy.

\begin{table}[h]
\vspace{-0.15in}
\centering
\resizebox{0.7\linewidth}{!}{
    \begin{tabular}{c|c}
     \hline
     Loss&$CD_{L2}\times100$\\
     \hline
     $L_{Pull}$&0.184\\
     $L_{Pull}+L_{\nabla}$&0.126\\
     $L_{Pull}+L_{\nabla}+L_{TV}$&0.056\\
     $L_{Pull}+L_{\nabla}+L_{TV}+L_{Surface}$&\textbf{0.040}\\
     \hline
   \end{tabular}}
    \vspace{0.05in}
   \caption{Effect of losses.}
   \label{table:ASLOSS}
   \vspace{-0.15in}
\end{table}

\begin{figure}[tb]
\vspace{-0.1in}
  \centering
   \includegraphics[width=\linewidth]{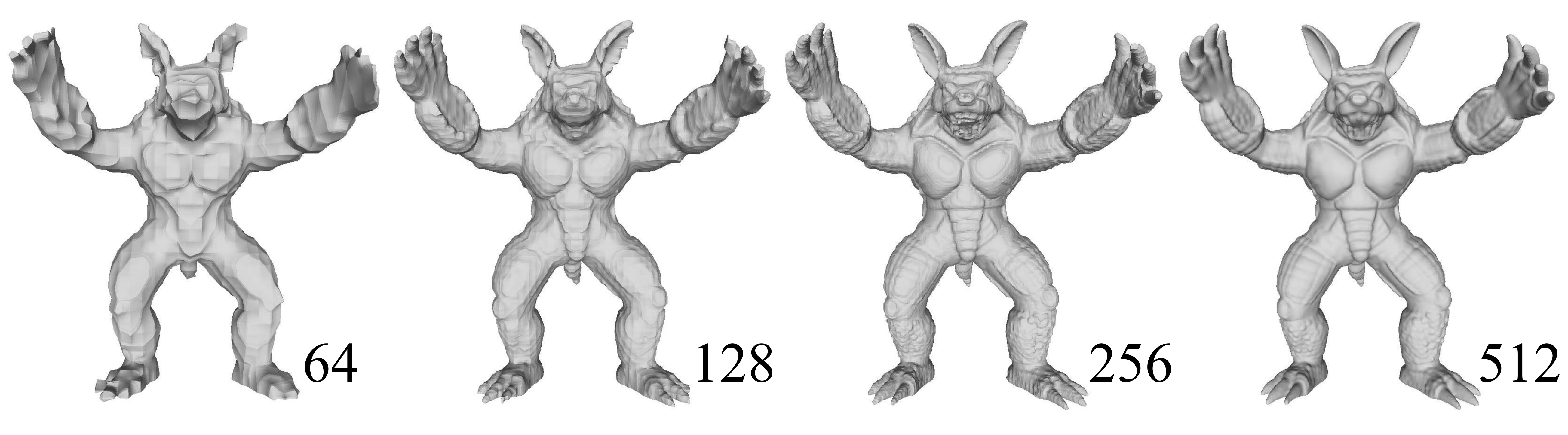}
  %
  %
  \vspace{-0.25in}
\caption{\label{fig:R}Effect of Resolution $R$.}
\vspace{-0.20in}
\end{figure}

\noindent\textbf{Resolution $R$. }We compare the effect of resolution $R$ on the reconstructed surfaces. We try several candidates $\{32,64,128,256,512\}$ to learn the discrete distance field. We use  the same bandwidth for the calculation of the pulling loss and the TV loss, and reconstruct meshes by running the marching cubes at the same resolution. The numerical and visual comparisons show that higher resolutions can infer more geometry details but also cost more inference time.

\begin{table}[h]
\vspace{-0.05in}
\centering
\resizebox{0.8\linewidth}{!}{
    \begin{tabular}{c|c|c|c|c|c}
     \hline
     &32&64&128&256&512\\
     \hline
     $CD_{L2}\times100$&0.186&0.097&0.048&0.040&0.035\\
     Time&50s&67s&84s&201s&645s\\
     \hline
   \end{tabular}}
    \vspace{0.05in}
   \caption{Effect of Resolution $R$.}
   \label{table:ASRESOLUTION}
   \vspace{-0.05in}
\end{table}

\noindent\textbf{Point Number. }We compare our performance on different numbers of points. We use the same set of shapes but with different numbers of points including $\{10K,100K,1000K\}$. The numerical comparison in Tab.~\ref{table:ABPOINTS} shows that our method can estimate a discrete distance field from different numbers of points well, which leads to more accurate reconstructed surfaces than the latest overfitting based methods. In addition, our time complexity is not affected by the number of points a lot, while the latest methods require more time to converge on more points.

\begin{table}[h]
\vspace{-0.0in}
\centering
\resizebox{0.8\linewidth}{!}{
    \begin{tabular}{c|c|c|c|c}  
     \hline
     Number&Method&$CD_{L2}\times 100$&Time&Memory\\
     \hline
     \multirow{3}{*}{10K}&NP&0.351&548s&4.5G\\
     &SAP&0.275&147s&3.5G\\
     &Ours&\textbf{0.122}&\textbf{82s}&\textbf{3.3G}\\
     \hline
     \multirow{3}{*}{100K}&NP&0.224&615s&4.6G\\
     &SAP&0.076&354s&3.6G\\
     &Ours&\textbf{0.040}&\textbf{84s}&\textbf{3.3G}\\
     \hline
     \multirow{3}{*}{1000K}&NP&0.175&1304s&4.8G\\
     &SAP&0.063&841s&3.7G\\
     &Ours&\textbf{0.039}&\textbf{90s}&\textbf{3.4G}\\
     \hline
   \end{tabular}}
   \vspace{0.05in}
   \caption{Effect of point numbers.}
   \label{table:ABPOINTS}
   \vspace{-0.10in}
\end{table}

\noindent\textbf{Initialization. }We highlight our geometric initialization of distances on grid vertices by comparing it with random initialization. The comparison in Tab.~\ref{table:INITIALIZATION} shows that random initialization makes the optimization converge hard, and cannot work well with the TV loss, especially with a bandwidth, which produces lots of artifacts in empty spaces. Moreover, we try an Eikonal constraint on gradients based on the geometric initialization, but did not get improvement.

\begin{table}[h]
\vspace{-0.05in}
\centering
\resizebox{0.8\linewidth}{!}{
    \begin{tabular}{c|c|c|c}
     \hline
     &Random&Geometric&$||\nabla||=1$\\
     \hline
     $CD_{L2}\times100$&0.155&\textbf{0.040}&0.054\\
     \hline
   \end{tabular}}
    \vspace{0.05in}
   \caption{Effect of Initialization.}
   \label{table:INITIALIZATION}
   \vspace{-0.05in}
\end{table}

\begin{figure}[tb]
  \centering
   \includegraphics[width=\linewidth]{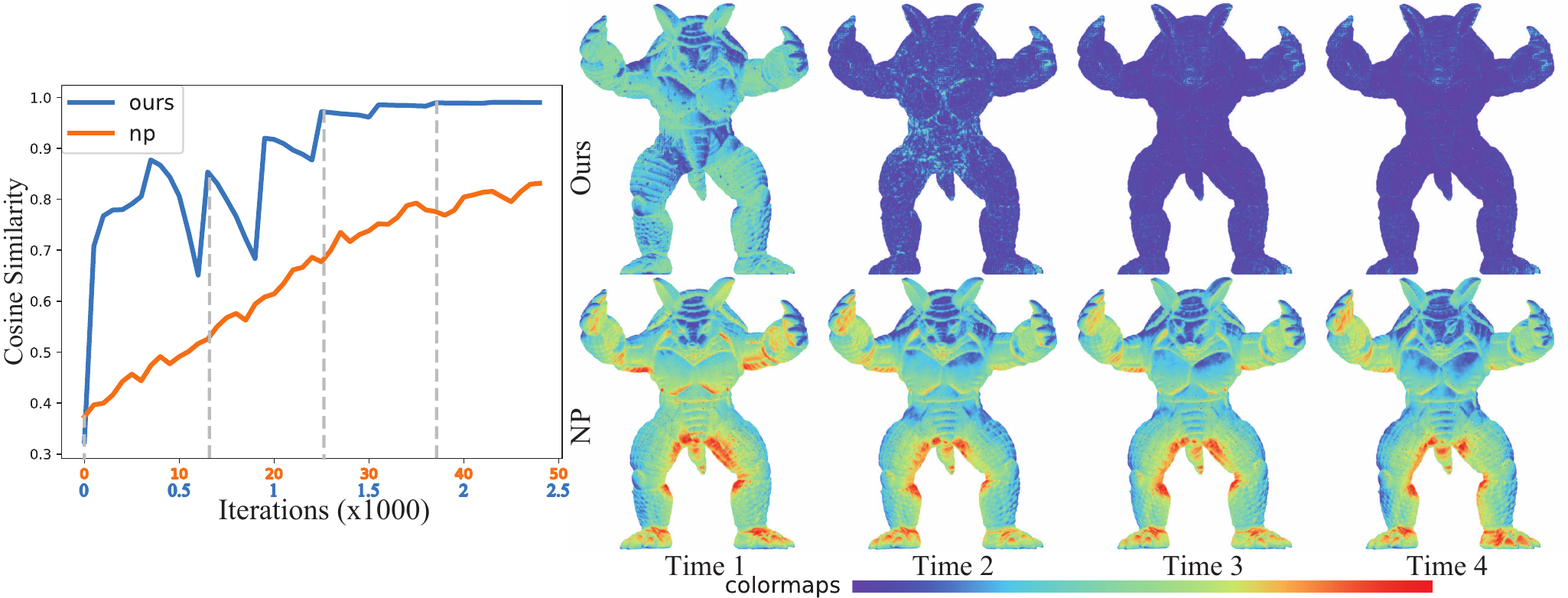}
  %
  %
  \vspace{-0.20in}
\caption{\label{fig:CompNP}Gradients comparison with neural networks.}
\vspace{-0.2in}
\end{figure}

\noindent\textbf{Bandwidth. }We report the effect of bandwidth $M_1$ and $M_2$ in the pulling loss and the TV loss in Tab.~\ref{table:ABBANDWIDTH}. For both losses, the bandwidth controls the scope of query sampling and continuous constraints. A larger bandwidth would cover more parameters which require more time and make the optimization converge slowly. For instance, if we sample queries and impose the pulling in all $R^3$ grids, the speed is even slower than NeuralPull which uses neural network for the pulling, since neural networks can generalize signed distances nearby and do not need to infer in all grids. Although a small bandwidth would save time, a too small bandwidth would cover too few grids around the surface, which leads to discontinuousness and artifacts. While if the bandwidth is large enough, we may not see surface improvement but just cost more time. Moreover, even with the same loss, NeuralPull does not produce better results than ours. Comparing with methods using neural networks, our discrete SDF can have faster convergence, as indicated by the angular error of gradients in Fig.~\ref{fig:CompNP}.

\begin{table}[h]
\vspace{-0.1in}
\centering
\resizebox{0.8\linewidth}{!}{
    \begin{tabular}{c|c|c}
     \hline
     Bandwidth&$CD_{L2}\times100$&Time\\
     \hline
     $L_{Pull}$ ($M_1=3$)&0.040&84s\\
     $L_{Pull}$ (All Grids)&0.043&628s\\
     $L_{TV}$ ($M_2=2$)&0.079&83s\\
     $L_{TV}$ ($M_2=14$)&0.040&84s\\
     $L_{TV}$ ($M_2=30$)&0.041&89s\\
     $L_{TV}$ ($M_2=50$)&0.040&96s\\
     $L_{TV}$ (All Grids)&0.040&101s\\
     NeuralPull&0.220&552s\\
     NeuralPull+$L_{\nabla}$+$L_{Surface}$&0.203&554s\\
     \hline
   \end{tabular}}
    \vspace{0.05in}
   \caption{Effect of Bandwidth.}
   \label{table:ABBANDWIDTH}
   \vspace{-0.1in}
\end{table}

\noindent\textbf{Noises. }We report our performance on point clouds with noise. We follow the experimental setting in PCP, and report our results with middle and large level noise in Tab.~\ref{table:ABNOISES}. To handle noise, we weight more on the continuous constraint $L_{TV}$ using a larger $\alpha=2$ to decrease the impact of noise. The comparison shows that our method can handle noise better than the neural network based methods. Since the continuousness of neural networks make gradients near the surface over smooth to overfit all noise, this makes it hard to infer accurate zero level-set.

\begin{table}[h]
\vspace{0.0in}
\centering
\resizebox{0.7\linewidth}{!}{
    \begin{tabular}{c|c|c|c}
     \hline
     Noise level&NP~\cite{Zhizhong2021icml}&PCP~\cite{DBLP:conf/cvpr/MaLZH22}&Ours\\
     \hline
     F-med-noise&0.280&0.071&\textbf{0.044}\\
     F-max-noise&0.310&0.298&\textbf{0.060}\\
     \hline
   \end{tabular}}
    \vspace{0.05in}
   \caption{Effect of noises.}
   \label{table:ABNOISES}
   \vspace{-0.1in}
\end{table}

\noindent\textbf{Normals. }We report our performance on point clouds with normals. If GT normal is available, we could add a supervision on normals at points on the surface. What we do is to supervise the gradients $\lambda f$ at points on the surface using their ground truth normals. The comparisons in Tab.~\ref{table:normal} show that the normal supervision improves the performance a little. This indicates that our method can infer pretty accurate gradients on the surface or near the surface, where gradients on the surface can be highly accurate to the normal ground truth on surfaces.

\begin{table}[h]
\vspace{-0.1in}
\centering
\resizebox{0.6\linewidth}{!}{
    \begin{tabular}{c|c|c}
     \hline
     Class&Ours&Ours(Normals)\\
     \hline
     $CD_{L2}$&\textbf{0.0401}&0.0403\\
     \hline
   \end{tabular}}
    \vspace{0.05in}
   \caption{Effect of normals.}
   \label{table:normal}
   \vspace{-0.2in}
\end{table}

\section{Conclusion}
We introduce GridPull for the fast inference of distance fields from large scale point clouds without using neural components. We infer a distance field using learnable parameters defined on discrete grids, which we can directly optimize in a more efficient way than neural networks. Besides the complexity advantage, our loss function manages to achieve more continuous distance fields with more consistent gradients, which leads to higher accuracy in surface reconstruction. We evaluate GridPull and justify its effectiveness on benchmarks. Numerical and visual comparisons show that GridPull outperforms the latest methods in terms of accuracy and complexity, even without using priors, normal, and neural networks.

{\small
\bibliographystyle{ieee_fullname}
\bibliography{papers}
}

\end{document}